# AGENT-BASED DECISION SUPPORT SYSTEM TO PREVENT AND MANAGE RISK SITUATIONS
# SYSTME D'AIDE À LA DÉCISION À BASE D'AGENTS POUR PRÉVENIR ET GÉRER DES SITUATIONS DE RISQUE

Kebair F. et Serin F.
Laboratoire d'Informatique de Traitement de l'Information et des Système
25 rue Philippe Lebon, 76058 Le Havre Cedex - France

**Résumé**
Le sujet de la prévention des risques et la réponse d'urgence est devenu une préoccupation politique et sociale d'envergure. L'une des approches proposées à ce défi est le développement des Systèmes d'Information et d'Aide à la Décision (SIAD) qui permettent d'aider les planificateurs d'urgence et les décideurs à détecter des risques de crises et de les gérer. Notre travail de recherche s'inscrit dans ce cadre, avec comme objectif le développement d'un SIAD qui doit être le plus générique possible.

**Summary**
The topic of risk prevention and emergency response has become a key social and political concern. One approach to address this challenge is to develop Decision Support Systems (DSS) that can help emergency planners and responders to detect emergencies, as well as to suggest possible course of actions to deal with the emergency. Our research work comes in this framework and aims to develop a DSS that must be generic as much as possible and independent from the case study.

## Introduction

In recent years we noticed a growth in natural and artificial disasters. This led the scientific community to be mobilized in order to find solutions to the issues related to the crisis management (also known as emergency response). Emergency response can be defined as the application of knowledge, procedures and activities to anticipate, prepare for, prevent, reduce or overcome any risk, harm or loss that may be associated with natural, technological or man-made crises and disasters during peacetime. Thereby, a decider needs necessarily a technical assistance that may help it to analyze in real time the situation and to make good decisions, before, during and after the crisis.

One of the technologies proposed so far, and that demonstrated their utilities and their efficiencies in this field are DSSs. To be successful such systems need to be adaptive, easy to use, robust and complete on important issues [7].

We intend through our research to bring our contribution in this domain by developing an information system that may help emergency managers and deciders to detect and to manage risk situations. We stress in our approach the characteristics of genericity and adaptivity of the system in order to construct a robust, flexible and efficient system that does not depend on a particular case and that may be adapted and used easily in new subject of studies. In order to validate our approach, we need to test the system on several applications in various domains. We are working currently on two applications addressed to risk and crisis management that are the game of Risk and the RoboCupRescue Simulation System (RCRSS) [6]. We choose these applications because in both we may have a complete knowledge of the environment and especially its components and the rules that define their behaviors and their interactions. For each application, an ontology of the domain is necessary in order to format information that describe the perceived environment. These information reflect spatial and temporal facts occurring in the environment. The role of the system is therefore to detect these facts, to analyze them and to find similar ones that have happened previously and for which we have a beforehand knowledge of their possible consequences and the appropriate actions to perform.

So far, a part of the system was implemented and tested [8][5]. This part intends to provide a dynamic representation of the current situation and its characterization. In this paper, we describe our approach in the devolvement of the system and its internal mechanism that we illustrate through the RCRSS case study.

## Risk, Crisis and Emergency Response

Risks are described as events which occurrences are uncertain and could have a negative impact [10], [11]. From a time perspective, an emergency is a serious situation or occurrence that happens unexpectedly and demands immediate action because of a potential threat to life or the environment. The urgent need for action or assistance to respond hints that the time-span is somewhat short; if no response is made the emergent situation will vanish and the result will obviously cause some kind of negative impact. The magnitude of an unattended emergency depends on who or what is affected.

A crisis is a turning point or decisive change in a crucial situation [1]. The uncertainty of the outcome is large; it can result in a disaster or pass almost unnoticed. The main identification of a crisis, as shown in Figure 1, is the uncertainty and it is characteristic by [12].

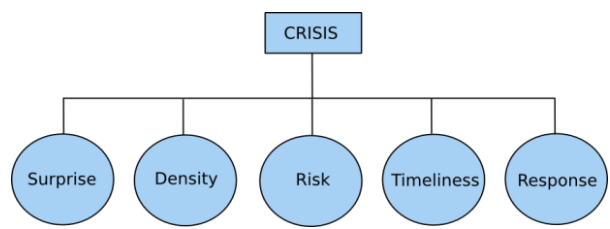

Figure 1. Crisis

Crisis may be described therefore as the manifestation of an unexpected risk that develops into a decisive period of acute difficulty [2]. Moreover, in every crisis situation a portion of emergency and disaster exists. Likewise, in emergencies, crisis situations arise to various degrees. The relation between emergency, disaster, and crisis is displayed together with the view that an emergency and a disaster need not contain surprise and response elements.

Emergency response can be defined as the application of knowledge, procedures and activities to anticipate, prepare for, prevent, reduce or overcome any risk, harm or loss that may be associated with natural, technological or man-made crises and disasters during peacetime.
The objectives of a response to an emergency are to minimize negative consequences (e.g., human and economic losses) and to ensure that the emergency will not escalate into disaster. However, events that trigger emergencies and the possibility of disaster are difficult to anticipate, due to many factors, among them the lack of information and the rapid change of the situation.

## DSS for Risk and Crisis Management

What is the purpose of DSS? The answer could be quite wide. In general, the purpose is "to improve the decision making ability of managers (and operating personnel) by allowing more or better decisions within the constraints of cognitive, time, and economic limits'" [4] More specifically, the purposes of a DSS are:

- Supplementing the decision maker
- Allowing better intelligence, design, or choice
- Facilitating problem solving
- Providing aid for non structured decisions
- Managing knowledge

Supplementing the decision maker dictates that decision support systems should supplement one or more of a decision maker's ability. In practice, this ability would involve analyzing, recognizing such as what happen if a fire starts in an oil refinery covering an area of ten hectares and employing 1000 of workers.
Based on Simon's decision processes, a DSS requires better intelligence, design, or choice by facilitate decision making phases [4]. For intelligence, a DSS should scan both internal organization and external environment. In the design phase, it could possibly generate and evaluate decision alternatives. DSS should take the form of offering advice to get maximize expected outcomes for the choice phase.
DSSs should help problem solving and make it more easy, smooth and fast. Also it should provide help for non structured decisions.
From the knowledge management, we could infer that DSSs should help to manage knowledge. It could be accomplished by strengthening knowledge representation and processing.
Thus, when all of the above purposes were understood, it might possible to get an idea what is the picture of putting together characteristics of DSS. We distinguish the five following characteristics:

- A DSS includes a body of knowledge: it describes some aspect of the decision maker's world, specifies how to accomplish various tasks, and indicates what conclusions are valid in various circumstances.
- A DSS has an ability to acquire and maintain descriptive knowledge and other kinds of knowledge.
- A DSS has ability to present knowledge in various customized ways as well as in standard reports.
- A DSS has an ability to select any desired subset of stored knowledge for either presentation or deriving new knowledge.
- A DSS can interact directly with a decision maker or a participant in a decision maker: it provides both a flexible choice and sequence if knowledge management activities.

What this DSS offers to users is to help deciders in their decision-making process in the case of a crisis or before a crisis occurs. Its main mechanism is based on the dynamic analysis of the current situation and its comparison with past situations. As it is shown in Figure 2, information about the current situation are sent by the actors of the environment, that may be also the end-users of the system. Information are thereafter treated by the kernel which is the main part of the system. Some knowledge about the specific domain are needed as the ontology of the domain that allows us to formalize entering information and the past situations that are permanently stored in a scenario base.

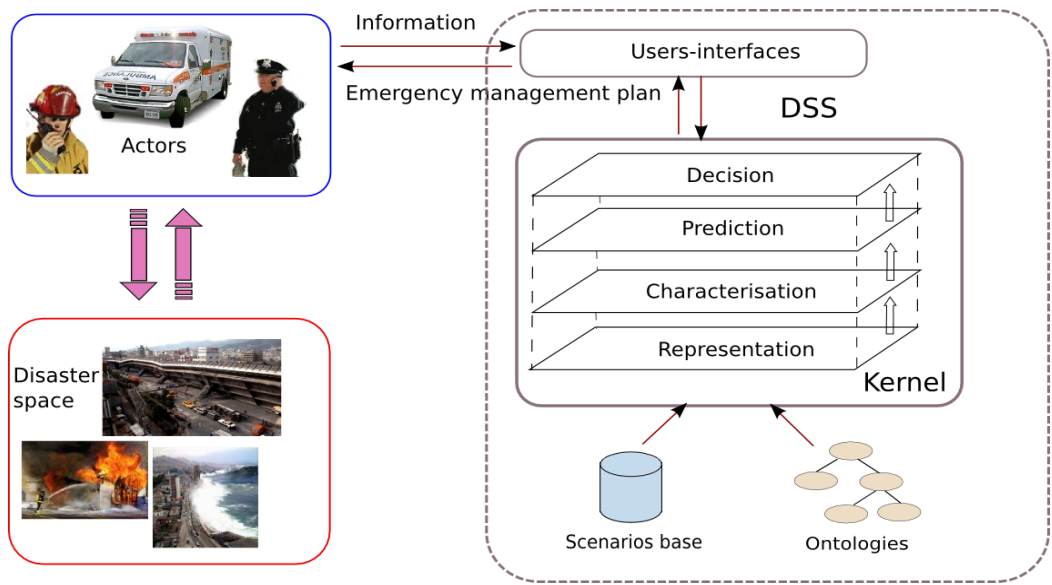

Figure 2. DSS architecture

The decision-support process is made up of four steps:
- The first step is to represent dynamically the current situation and its evolution thanks to information perceived from the environment;
- The second step is to characterize the situation in order to extract the significant facts of the situation;
- The third step is to evaluate the situation by comparing it with past situations;
- The fourth step is to provide the final decision to users basing-on the result of the previous step.

## Case of Study: RoboCupRescue Simulation System

**RoboCupRescue**

RCRSS intends to simulate an earthquake scenario including various kinds of incidents as the traffic after earthquake, buried civilians, road blockage and fire accidents. A set of heterogeneous agents coexist in the disaster space, each with a specific goal and a particular role. Our purpose here is to enhance the decision-making of the RCR agents in their rescue operations thanks to the DSS.

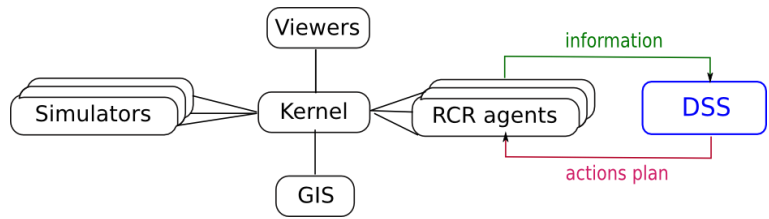

Figure 3. RCRSS architecture

Information received from the environment are treated by a multilayer agent-oriented process before providing the final result to deciders. In the following sections, we describe and explain this process while taking as example of application RCRSS and while showing some experimentations that we could test so far.

**Information formalization**

The first step of our approach is to format information coming from the environment in the shape of Factual Semantic Features (FSF). For that, we need to reify the observations issued from the perceived environment. In previous works, we presented an object-model that categorizes these observations [5]. Basing-on this object-model we define then the ontology of the domain that allows us to format information. In the context of sharing knowledge, "the term ontology means a specification of a conceptualization, a description (like a formal specification of a program) of the concepts and relationships that can exist for an agent or a community of agents." [3].

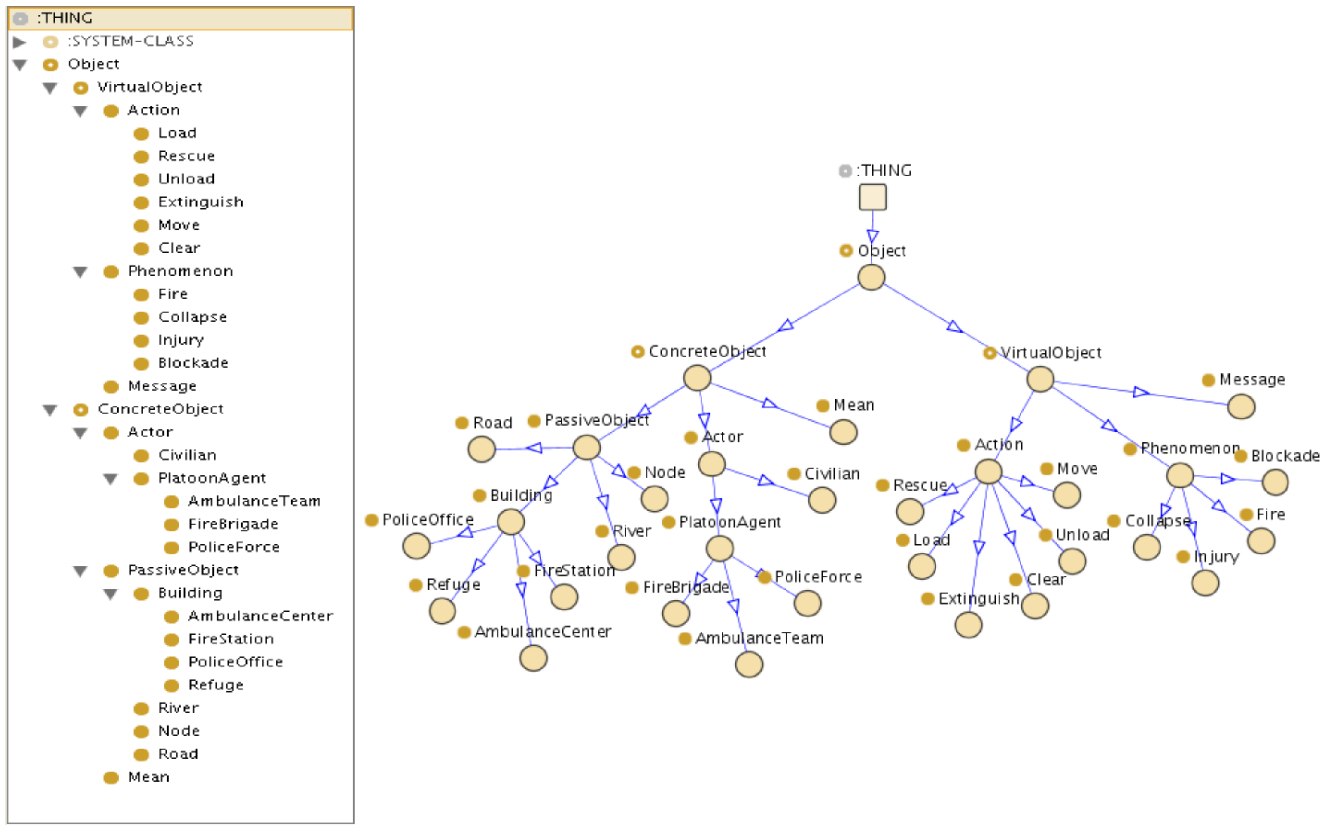

Figure 4. RoboCupRescue ontology

Each FSF describes a fact occurred in the environment and is managed by a factual agent inside the system. We mean by "agent" hardware or (more usually) software-based computer system that operates autonomously, sometimes for a specific goal, and interacts with other agents [13]. An FSF has the following structure: <selector, (qualifier, value) $^+$>. The selector is related to a perceived object of the environment to which are joined some characteristics described by qualifiers and the values of these qualifiers. An example of an FSF related to a fire is: (fire#14, fieriness, 1, inDangerNeighbours, 3, burningNeighbours, 2, localisation, 20|25, time, 7). This FSF means a fire is ignited in building number 14 with intensity equals to 1, has 2 burning fires and can spread to 3 neighbour buildings. The burning building has the following coordinates 20|25, and the fire was perceived in the 7$^{th}$ cycle[1] of the simulation.

Figure 4 shows the ontology of the RoboCupRescue (RCR) environment:

− Concrete objects: gather essentially RCR agents (civilian and rescue agents), buildings and roads.
− Virtual objects: include the incidents that can happen in the RCR environment as fires, roads blockade, civilian injuries and building collapses. They include also the actions performed by the actors and the messages exchanged between them.

FSFs may have a relation with each other. This relation is founded according to a semantic, temporal and spatial aspect. We must define therefore a proximity measure in order to evaluate this relation to know consequently what is the relation between two FSFs. Thereby, two FSFs may be opposite (close) if they have a negative (positive) proximity or neutral if the proximity is null.

**Situation representation and characterisation**

The set of factual agents forms a representation MultiAgent System (MAS) which composes the upper layer of the system. The goal of this layer is to provide dynamically and in real time a representation of the current situation. This representation is obtained thanks to factual agents interactions; from the one hand alliances are constituted containing agents having positive proximity measures and on the other hand oppositions are created between agents having negative proximity measures.

So far, we started tests on the upper layer by representing fires incidents in RCR. In the table of Figure 5 a part of the internal state of the representation MAS is shown. Two factual agents: fire brigades (in red color and starting with « fb ») and fires (in orange color and starting with « f ») are displayed. Each factual agent has a generic automaton, composed by four states, that manages its behavior and two indicators that reflect its dynamics. The evolution of these indicators is displayed with two horizontal bars: red and blue colors represent positive values and purple color represents negative values. Both fire brigades and fires factual agents are strong when they are in the third state of the automaton and in which they have high values of their indicators. In this state, fire brigades are close to fires and try to extinguish them by forming groups as it is shown in the virtual disaster space of RCRSS in Figure 5. Fires are also important in this state, since they are at their beginning, which means they can spread to their neighborhood, but also they are easy to extinguish and represent consequently priority targets to extinguish for fire brigades.

Factual agents are characterized during their evolution. In other words, the system refines the current representation by extracting the salient facts of the situation. This characterization is based on the analysis of factual agents dynamics and more precisely the evolution of their internal indicators. This process leads to the formation of factual agents groups and is ensured by the characterization layer of the kernel where each group is managed by a characterization agent. We distinguish three factual agents groups:

− Active agents: including agents of which at least one of their indicators values increased;
− Passive agents: including agents of which the two indicators values decreased;
− Stable agents: including agents of which indicators values did not change.

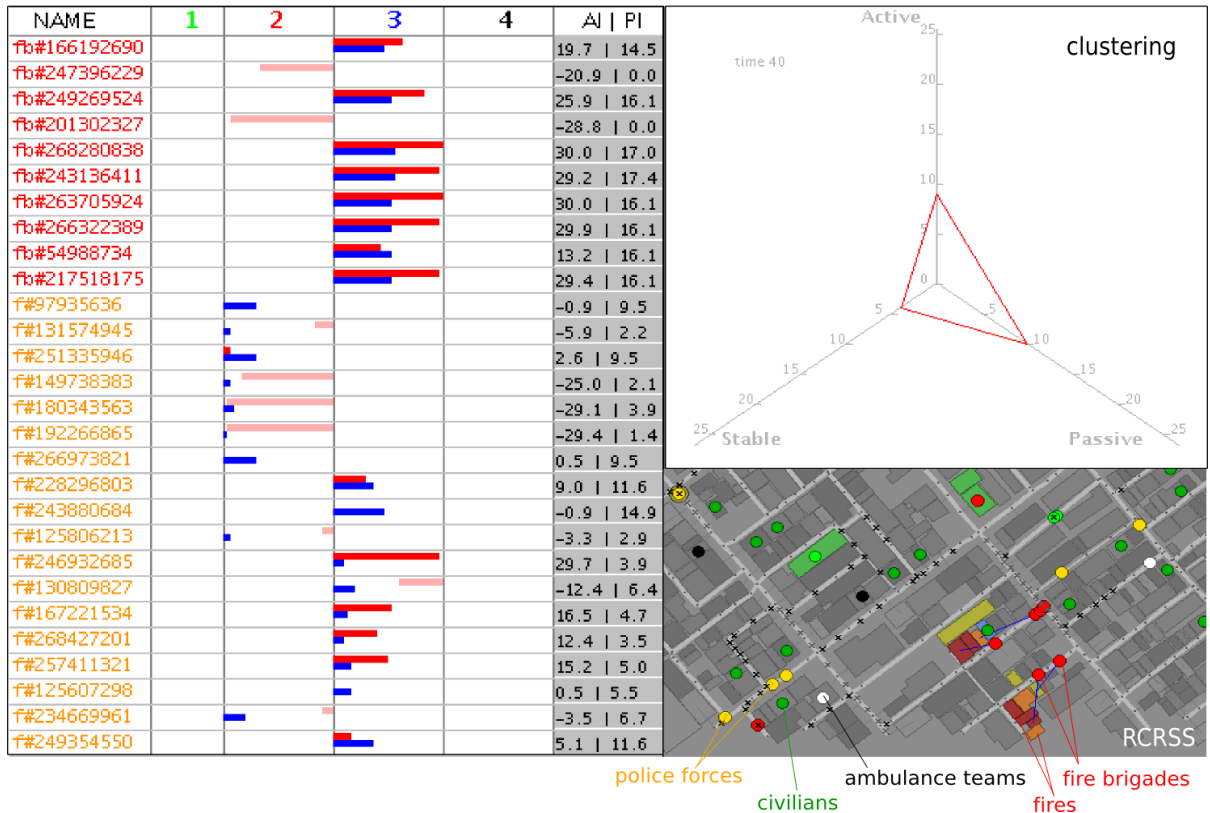

Figure 5. Graphic tools to visualize and analyze the representation MAS

---

1 One cycle in the simulation equals one second

Basing-on this classification, the system detects the changes occurred in the situation and evaluates them. Thus, in the case of fires here, the system may read into Active agents group new incidents discovery or an encounter between fire brigades and fires. Passive agents may represent on extinction fires or fire brigades which are in difficulty. Finally, Stable agents may represent fire brigades and fires which are isolated or for which the system did not receive further information. Consequently, these agents do not change generally and none reaction is therefore noticed.

A case-based reasoning is used in order to perform this evaluation where each observed situation is compared with a past situation. Each evaluation provides the possible consequences of the situation and the actions that can be made. Moreover, the system must learn over time by improving the suggested decisions by considering weight the results of these decisions. This step is managed by prediction agents of the third layer of the kernel and of which design and implementation are ongoing.

## **Conclusion**

In this paper we propose an agent-oriented approach for a DSS that can be used to improve decision-making in emergency and risk situations. The work accomplished until now concerns a part of the system that allows the representation and the characterization of the current situation. We choose the RCRSS in order to test the two first steps of our approach. This allowed us to validate the environment modeling and the formalization of its extracted information, as much as to study the behavior of the representation MAS and its characterization. We intend currently to define several scenarios and to set-up a case-based reasoning for the prediction layer in order to finalize the process by evaluating the observed situation and providing decisions to the rescue agents.

**Références**

[1] Asian Disaster Reduction Center, Total Disaster Risk Management - Good Practices, Kobe, Japan, January 2005.
[2] Fink, S., Crisis Management: Planning for the Inevitable, Amacom, New York, 1986.
[3] Gruber, T. R., A translation approach to portable ontologies. Knowledge Acquisition', 5(2):199-220, 1993.
[4] Holsapple, C.W., Whinston, A.B., Decision Support Systems: A Knowledge-based Approach, West Publishing Company, Minneapolis/St Paul, 713 pages, 1996.
[5] Kebair, F., Serin, F. and Bertelle, C., Agent-Based Perception of an Environment in an Emergency Situation, Proceedings of The 2007 International Conference of Computational Intelligence and Intelligent Systems (ICCIIS), World Congress of Engineering (WCE), 1999, London, U.K., pp. 49-54.
[6] Kitano, H., Tadokor, S., Noda, H., Matsubara, I., Takhasi, T., Shinjou, A. and Shimada, S., Robocup-rescue: Search and rescue for large scale disasters as a domain for multi-agent research, Proceedings of the IEEE Conference on Systems, Man, and Cybernetics (SMC-99), 1999, vol. 6, pp. 739-743.
[7] Little, J.D.C., Models and managers: The concept of a decision calculus, Management Science, n° 8, vol. 16, pp. 466-485, 1970.
[8] Person, P. Boukachour, H., Coletta, M., Galinho, T. and Serin, F., Data representation layer in a MultiAgent decision support system, IOS Press, vol. 2, no. 2, pp. 223-235, 2006.
[9] Rinaldi, S.M., Peerenboom J.P., and Kelly T.K., Complexities in Identifying, Understanding, and Analyzing Critical Infrastructure Interdependencies, 2001, IEEE Control Systems Magazine: 11-25.
[10] Wallace, L., Keil, M., Software Project Risks and Their E_ect on Outcomes, Communications of the ACM, 2004, vol. 47, no. 4, 68- 73.
[11] Webster, K.P.B., de Oliveira, K.M., Anquetil, N., A Risk Taxonomy Proposal for Software Maintenance, ICSM05. Proceedings of the 21st IEEE International Conference, 2005, 453-461.
[12] Will, D., Definition of crisis, http://www.lrz-muenchen.de/~ua352bm/webserver/webdata/Will/node2.html, last accessed May 25, 2008.
[13] Wooldridge, M., Jennings, N.R., Intelligent agents: Theory and practice, The Knowledge Engineering Review, 10(2):115152, 1995.